\documentclass{article}
\usepackage{spconf,amsmath,epsfig}
\usepackage{pbox}
\usepackage{tikz}
\usepackage{pgfgantt}
\usepackage{subfigure}
\usepackage{gensymb}
\usepackage{amsmath}
\usepackage{adjustbox} 
\usepackage{amsfonts}

\usepackage{array}
\usepackage{graphicx}
\usepackage{multirow}

\usepackage{soul}

\usetikzlibrary{matrix,shapes,arrows,positioning,chains}
\tikzstyle{process} = [rectangle, minimum width=3cm, minimum height=1cm, text centered, draw=black]
\tikzstyle{arrow} = [thick,->,>=stealth]
\tikzstyle{io} = [trapezium, trapezium left angle=70, trapezium right angle=110, text centered]
\tikzstyle{decision} = [diamond, draw, text  centered, inner sep=1pt]

\usepackage{times}
\usepackage{epsfig}
\usepackage{graphicx}
\usepackage{amsmath}
\usepackage{amssymb}
\usepackage{pifont}
\usepackage[ruled,vlined]{algorithm2e}
\SetArgSty{textnormal}

\let\oldnl\nl
\newcommand{\nonl}{\renewcommand{\nl}{\let\nl\oldnl}}
\usepackage{makecell}
\usepackage{booktabs} 
\usepackage{multirow}
\usepackage{color, xcolor}
\usepackage{tabularx,verbatim}
\usepackage{xspace}

\usepackage[labelsep=period]{caption}

\usepackage{setspace}
\usepackage{comment}
\usepackage{color, colortbl}

\usepackage{amsmath,epsfig}
\usepackage{pbox}
\usepackage{tikz}
\usepackage{pgfgantt}
\usepackage{subfigure}
\usepackage{gensymb}
\usepackage{amsmath}
\usepackage{adjustbox} 
\usepackage{amsfonts}

\usepackage{array}
\usepackage{graphicx}
\usepackage{multirow}

\usetikzlibrary{matrix,shapes,arrows,positioning,chains,plotmarks}
\tikzstyle{process} = [rectangle, minimum width=3cm, minimum height=1cm, text centered, draw=black]
\tikzstyle{arrow} = [thick,->,>=stealth]
\tikzstyle{io} = [trapezium, trapezium left angle=70, trapezium right angle=110, text centered]

\definecolor{Gray}{gray}{0.92}
\newcolumntype{L}{>{\arraybackslash}m{2cm}}
\DeclareMathOperator*{\Max}{max}
\DeclareMathOperator*{\ArgMax}{argmax}
\DeclareMathOperator*{\Argmin}{argmin}
\DeclareMathOperator*{\Min}{min}

\title{Reiterative Domain Aware Multi-Target Adaptation}
\name{Sudipan Saha$^{1}$, Shan Zhao$^{1}$, Nasrullah Sheikh$^{2}$, Xiao Xiang Zhu$^{1,3}$}

\address{Technical University of Munich, Taufkirchen/Ottobrunn, Germany$^{1}$ \\ IBM Research Almaden, CA, USA $^{2}$ \\  German Aerospace Center (DLR),  We{\ss}ling, Germany$^{3}$}

\begin{document}

%
\maketitle
\begin{abstract}
Most domain adaptation methods focus on single-source-single-target adaptation settings. 
Multi-target domain adaptation is a powerful extension in which a single classifier is learned for multiple unlabeled target domains.
To build a multi-target classifier, it is important to have: a feature extractor that generalizes well across domains; and effective aggregation of features from the labeled source and different unlabeled target domains.
Towards the first, we use the recently popular  Transformer as a feature extraction backbone. Towards the second, we use a co-teaching-based approach using a dual-classifier head, one of which is based on the graph neural network. The proposed approach uses a sequential adaptation strategy that adapts one domain at a time starting from the target domains that are more similar to the source, assuming that the network finds it easier to adapt to such target domains.
After adapting on each target, samples with a softmax-based confidence score greater than a threshold are added to the pseudo-source, thus aggregating knowledge from different domains. However, softmax is not entirely trustworthy as a confidence score and may generate a high score for unreliable samples if trained for many iterations.  To mitigate this effect, we adopt a reiterative approach, where we reduce target adaptation iterations, however, reiterate multiple times over the target domains. The experimental evaluation on the Office-Home, Office-31 and DomainNet datasets
shows significant improvement over the existing methods. We have achieved 10.7$\%$ average improvement in Office-Home dataset over the state-of-art methods. 
\end{abstract}
%

\fboxsep=0mm
\fboxrule=0.1pt

\section{Introduction}
\label{sectionIntroduction}
The deep learning techniques have produced an impressive performance for a wide array of visual inference tasks \cite{lecun2015deep,goodfellow2016deep,ronneberger2015u,saha2019semantic,guo2017deep}. Yet,  these models fail to generalize when exposed to a new environment. This is caused by a misalignment between the source and the target data distributions, causing the trained model to perform poorly during inference \cite{zhang2019bridging}. Since collecting labeled data for each domain is challenging, a rich line of research, unsupervised domain adaptation \cite{long2016unsupervised,ganin2015unsupervised}, has evolved to effectively exploit the source data to learn a robust classifier on the target domain(s). 
\par

Most domain adaptation methods are designed to adapt to a single unlabeled target from a single labeled source domain.  Such methods include those based on generative modeling \cite{hong2018conditional,bousmalis2017unsupervised}, adversarial training \cite{tzeng2017adversarial}, and statistical alignment \cite{long2014transfer,peng2019moment,li2018adaptive}. However, such models are not suitable for practical settings where we may come across many target domains as a separate model needs to be trained for each target domain. Recently, some works in the literature have addressed this issue by designing methods to adapt to multiple domains simultaneously from a single source domain \cite{nguyen2021unsupervised}. This domain adaptation setting is called Multi-target Domain Adaptation (MTDA). MTDA methods can be both domain-agnostic/aware, depending on the availability of the domain labels of target samples \cite{roy2021curriculum}. 
\par
MTDA is challenging as the increase of the domains brings more difficulty in aligning feature distributions among them. While previous works focus on merely aligning cross-domain distributions \cite{roy2021curriculum}, the generalization ability of the backbone feature extractor may play a crucial role
in multi-target adaptation. Transformers have recently shown excellent capability in many computer vision tasks \cite{han2020survey}. Unlike  CNN  which generally concentrates on the local features, the Vision Transformer (ViT) \cite{dosovitskiy2020image} exploits the attention across the patches to capture the long-distance features and acquires global information.  Moreover, Transformer-based models have shown good performance in transfer learning and domain adaptation \cite{malpure2021investigating,duong2021detection,yang2021transformer}. Motivated by this, we propose to use pre-trained Transformer as a feature extractor for MTDA. We hypothesize that the strong generalization capability of Transformer-based features will enhance the adaptation ability across multiple target domains.  
\par
Furthermore, to learn an effective multi-target classifier, it may be useful to aggregate features from different domains. To this end, graph neural networks (GNNs) have been found to be effective  \cite{zhou2020graph}. By exploiting GNN's potential in tandem with co-teaching and curriculum learning, Roy \textit{et. al.} \cite{roy2021curriculum} proposed Domain-aware Curriculum Graph Co-Teaching (D-CGCT) for MTDA.  Curriculum learning is a type of learning in which easy examples are tackled first while gradually increasing the task difficulty \cite{jiang2015self,graves2017automated}. Inspired by this, we adopt this paradigm in conjunction with Transformer-based feature extraction. 
Similar to  \cite{roy2021curriculum}, we use a 
dual classifier comprising of a MLP head and a GNN head.  One target domain is processed at a time, starting from the easiest domain or the domain closest to the source domain. Minibatches comprising of  source samples and target samples are fed to the dual-head classifier. While the MLP based classifier focuses on the individual samples, the GNN based classifier aggregates features from different samples in the minibatch. Similar to co-teaching \cite{han2018co}, they help each
other in an iterative manner to learn effective target classifier. The pseudo-labels are selected after processing each domain and are used along with the source domain samples for processing the following target domains. The crux of the idea is that pseudo-labeled samples (henceforth called pseudo-samples) from the easier domains assist in adapting the classifier on the more difficult target domains.
One caveat of this idea is that it is  not trivial to select pseudo-samples from the target domains. In \cite{roy2021curriculum}, a simple strategy based on softmax score is used. After adapting on a target domain, the target samples exceeding a threshold softmax value are selected as pseudo-labeled samples.
However, softmax is not a reliable confidence indicator for
deep learning models \cite{guo2017calibration,hein2019relu} and may yield overconfident
predictions if trained for the high number of training iterations. Reducing the number of training iterations over a target domain may lead to the selection of the higher quality, however lesser number of pseudo samples. To compensate lesser pseudo samples selected in one training cycle,  we postulate that the MLP-GNN dual classifier can be better adapted to target domains by reiterating multiple times over the set of target domains. Pseudo-source is initialized with the source samples and ingests some target pseudo-samples after each reiteration. In practice, the total number of training iterations is conserved as in D-CGCT. 
\par
In contrast to D-CGCT \cite{roy2021curriculum}, we use more source samples than target samples in each minibatch during training. Even though our goal is to adapt the classifier on the target domains, the primary reference of guidance for the classifier is the labeled source samples. Hence we postulate that more source samples in comparison to the target samples in a minibatch can improve the adaptation performance.
\par
The contributions of this work are as follows:
\begin{enumerate}
    \item We introduce Transformer in the context of multi-target adaptation.
    \item We combine Transformer and GNN in the same framework to obtain a superior MTDA framework that benefits simultaneously from the excellent generalization ability of the Transformer backbone and feature aggregation capability of GNN. 
     \item We furthermore show that by choosing more source samples than target samples in the training minibatch improves MTDA performance.

    \item We experimentally validate on three popular benchmark datasets showing a significant improvement over the existing domain-aware MTDA frameworks, e.g.,  10.7$\%$ average improvement over D-CGCT in Office-Home dataset.
\end{enumerate}

\section{Related Works}
\label{sectionRelatedWork}

Mitigating the discrepancy between the training and test data distributions is a long lasting problem.
The machine learning and computer vision literature is rich in domain adaptation techniques for several problems including image classification, semantic segmentation, and object recognition \cite{zhang2019bridging}. 
In this regard, domain adaptation aligns the data distribution using generative modeling \cite{hong2018conditional,bousmalis2017unsupervised}, adversarial 
training \cite{tzeng2017adversarial,long2017conditional}, and standard divergence measures \cite{peng2019moment,li2018adaptive}. The domain adaptation problem can be presented in various flavors depending on the overlap of
classes between domains (closed-set, open-set) \cite{panareda2017open,saito2018open,jing2021towards} or number of sources and targets, e.g., single-source-single-target, multi-source \cite{sun2015survey,zhang2015multi,gong2021mdalu}, multi-target \cite{yu2018multi,isobe2021multi,gholami2020unsupervised}.
\par
Multi-target domain adaptation \cite{roy2021curriculum,yu2018multi,yang2020heterogeneous,isobe2021multi,chen2019blending,nguyen2021unsupervised} transfers a network learned on a single labeled source dataset to multiple unlabeled target
datasets. This line of research is quite new and there are only few methods towards this direction. While there has been extensive research on single-target domain adaptation, those methods cannot be trivially applied 
in the multi-target setting.  The method in \cite{chen2019blending} clusters  the blended
target domain samples to obtain sub-targets and subsequently applies a single-target domain adaptation method
on the source and the obtained sub-targets. Multi-teacher MTDA (MT-MTDA) \cite{nguyen2021unsupervised} uses knowledge
distillation to distill target domain
knowledge from multiple teachers to a common student.
\par
Sequentially adapting the source classifier to each target domain for multi-target adaptation is related to incremental learning. Incremental learning \cite{rosenfeld2018incremental} continually
trains a classifier for sequentially available data for novel tasks. There are only few existing works in the domain adaptation literature \cite{mancini2019adagraph,wulfmeier2018incremental} in this line.
\par
Learning from noisy labels is an important topic in weakly supervised learning. A recently popular approach to handle noisy labels is co-teaching \cite{han2018co} that cross-trains two deep networks that teach each other given every minibatch.
While the predictions from the two classifiers may strongly disagree at the beginning, the two networks converge towards consensus with the increase of training epochs. 
\par
Graph Neural Networks (GNNs) are neural networks applied on the graph-structured data. GNNs can capture the relationships between
 the nodes (objects) in a graph using the edges \cite{kipf2016semi}. GNNs \cite{kipf2016semi} have been recently
 adopted for progressive/incremental domain adaptation \cite{luo2020progressive,roy2021curriculum}.
Progressive Graph Learning (PGL) \cite{luo2020progressive}  operates in single-target setting where a graph neural network with episodic training is integrated to suppress the underlying conditional shift and to close the gap between the source
 and target distributions. Further extending this concept, \cite{roy2021curriculum} proposes a method for multi-target domain adaptation. As discussed in Section \ref{sectionIntroduction}, our method is closely related to
\cite{roy2021curriculum}. While their method is applicable
to both domain-agnostic and domain-aware setting, in our work we focus on the
domain-aware setting in \cite{roy2021curriculum}, i.e., D-CGCT. A heterogeneous Graph Attention
Network (HGAN) is used for MTDA in \cite{yang2020heterogeneous}.
\par
Transformer is a recently popular deep learning architecture. The fundamental building block of a Transformer is self-attention. Attention mechanism computes the responses at each token in a sequence by attending it to all tokens and gathering the corresponding embeddings based on the attention scores accordingly \cite{han2020survey}. To extend the application of Transformer to the grid-like data, ViT is proposed, which uses a pure Transformer and formulates the classification problem as a sequential problem by splitting the image into a sequence of patches. It is composed of three main components: a linear layer to project flattened image patches to lower dimensional embeddings, the standard Transformer encoder with multi-head self-attention to compute independent scoring functions from different subspaces, and a MLP head for classification score prediction. Different variants of image-based have been proposed, e.g., Data-efficient image Transformers (DeiT) \cite{touvron2021training}, Convolutional vision Transformer(CvT)\cite{wu2021cvt}, Compact Convolutional Transformer(CCT) \cite{hassani2021escaping}, and LocalViT \cite{li2021localvit}. Recently the application of Transformers in computer vision tasks including image generation \cite{lee2021vitgan}, object detection \cite{beal2020toward}, classification and segmentation \cite{wu2021fully} has sprung up due to its advantageous capability. The Transformer encoder in ViTs can be regarded as a feature extractor naturally. Unlike CNN which merely concentrates on local features, the attention across the patches captures the long distance features and acquires global information.  Transformers have shown good performance in transfer learning \cite{malpure2021investigating}, e.g. in medical image analysis \cite{duong2021detection}. Furthermore, Transformers have been shown to be useful for source-free domain adaptation in  \cite{yang2021transformer}. They make the observation that focusing attention on the objects in the image plays an important role in domain adaptation.
\par
Similar to \cite{yang2021transformer}, our work exploits Transformer as backbone feature extractor to capitalize on its ability to generalize across domains. MTDA is a relatively new area, where our work is related to \cite{roy2021curriculum}. However, in contrast to \cite{roy2021curriculum}, we use a reiterative approach to effectively ingest target domain pseudo-samples. The learning of the dual-classifier head is based on co-teaching \cite{han2018co} and follows similar episodic training as in \cite{luo2020progressive}.

\begin{table}[t]
\centering
\begin{tabular}{l|l|l} 
 \hline
\textbf{Network component} & \textbf{Architecture} & \textbf{Output}  \\ 
\hline
Feature extractor   & Transformer & $B\times256$  \\
($F$) & with bottleneck &    \\
\hline
MLP classifier   & FC layer & $B\times n_c$ \\
($G_{mlp}$) & &    \\
\hline
Edge network & Conv(256,256,1), & $B\times B$ \\
($f_{edge}$) & Conv(256,128,1), & \\
 & Conv(128,1,1) & \\
\hline
Node classifier & Conv(512, $2\times n_c$,1), & $B\times n_c$ \\
($f_{node}$) & Conv($2\times n_c$,$n_c$,1) & \\
\end{tabular}
\caption{Network architecture assuming batch-size $B$. Conv(a,b,c) denote a convolutional filter with $a$ input features, $b$ output features, and  kernel size $c\times c$.}
\label{tableNetworkArchitecture}
\end{table}

\tikzset{
    between/.style args={#1 and #2}{
         at = ($(#1)!0.5!(#2)$)
    }
}

\begin{figure}
 \centering
 \begin{tikzpicture}
\node (1a) [draw, io, align=center] {Source $\mathcal{S}$};
\node (1b) [draw, io, align=center, right of=1a,xshift=3cm] {Target $\mathcal{T}_j$};
\node (2) [rounded corners=3pt,draw, align=center, between=1a and 1b, yshift=-1.10cm] {Transformer-based feature extractor};
\node (3a) [rounded corners=3pt,draw, align=center, below of=2, xshift=-3.85cm, yshift=-0.50cm] {MLP\\ classifier};
\node (3b) [rounded corners=3pt,draw, align=center, below of=2, yshift=-0.50cm] {Edge\\ classifier};
\node (3c) [rounded corners=3pt,draw, align=center, below of=2, xshift=+1.85cm, yshift=-0.50cm] {Domain\\ discriminator};

\node (4) [rounded corners=3pt,draw, align=center, below of=3b, yshift=-0.50cm] {Node\\ classifer};

\node (5) [rounded corners=3pt,draw, align=center, below of=4, yshift=-0.50cm] {Update pseudo-source};

\draw [arrow] (1a) -- (2);
\draw [arrow] (1b) -- (2);
\draw [arrow] (2) -- (3a);
\draw [arrow] (2) -- (3b);
\draw [arrow] (2) -- (3c);
\draw [>=triangle 45, ->] (3a) -- (3b);
\draw [arrow] (3b) -- (4);
\draw [arrow] (4) -- (5);
\draw [arrow] (5.west) -| (3a.south);

\node (ancillary1) [between=3a and 3b, yshift=-0.3cm] {Pseudo-labels};
\end{tikzpicture}
 \caption{Proposed method for adapting on a single target domain $\mathcal{T}_j$.}
 \label{figureSingleTargetAdaptation}
\end{figure}
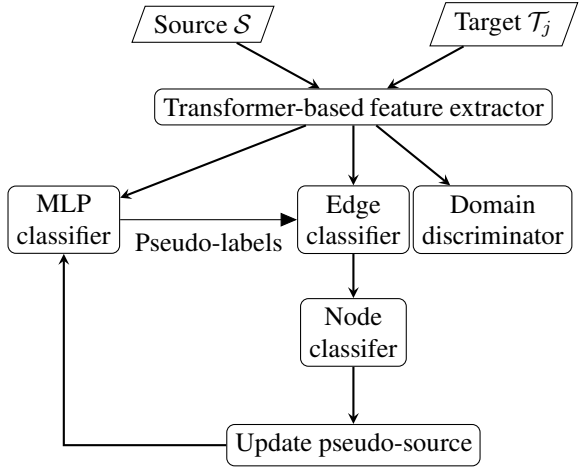

\tikzset{
    between/.style args={#1 and #2}{
         at = ($(#1)!0.5!(#2)$)
    }
}
\begin{figure}[h]
 \centering
 \begin{tikzpicture}
  \node (0a) [draw, io, align=center] { Source $\mathcal{S}$};
  \node (0b) [draw, align=center,right of=0a,xshift=2cm] {Source training};
  \node (0c) [draw, io, align=center,right of=0b,xshift=2cm] { Model};

 \node (1a) [draw, io, align=center,below of=0a,yshift=-0.40cm] { $\mathcal{S}$+$\mathcal{T}_1$};
  \node (1b) [draw, align=center,right of=1a,xshift=2cm] {Target adaptation};
  \node (1c) [draw, io, align=center,right of=1b,xshift=2cm] { Model, $\hat{\mathcal{S}}^{total}$};
  
\node (2a) [io,draw, align=center, below of=1a,yshift=-0.40cm]{$\hat{\mathcal{S}}^{total}$+ $\mathcal{T}_2$}; 
\node (2b) [draw, align=center,right of=2a,xshift=2cm] {Target adaptation};
  \node (2c) [draw, io, align=center,right of=2b,xshift=2cm] { Model, $\hat{\mathcal{S}}^{total}$};

\node (3a) [io,draw, align=center, below of=2a,yshift=-0.40cm]{$\hat{\mathcal{S}}^{total}$+ $\mathcal{T}_3$}; 
\node (3b) [draw, align=center,right of=3a,xshift=2cm] {Target adaptation};
  \node (3c) [draw, io, align=center,right of=3b,xshift=2cm] { Model, $\hat{\mathcal{S}}^{total}$};

\node (4a) [io,draw, align=center, below of=3a,yshift=-0.40cm]{$\hat{\mathcal{S}}^{total}$+ $\mathcal{T}_1$}; 
\node (4b) [draw, align=center,right of=4a,xshift=2cm] {Target adaptation};
  \node (4c) [draw, io, align=center,right of=4b,xshift=2cm] { Model, $\hat{\mathcal{S}}^{total}$};
  
  \node (5a) [io,draw, align=center, below of=4a,yshift=-0.40cm]{$\hat{\mathcal{S}}^{total}$+ $\mathcal{T}_2$}; 
\node (5b) [draw, align=center,right of=5a,xshift=2cm] {Target adaptation};
  \node (5c) [draw, io, align=center,right of=5b,xshift=2cm] { Model, $\hat{\mathcal{S}}^{total}$};

\node (6a) [io,draw, align=center, below of=5a,yshift=-0.40cm]{$\hat{\mathcal{S}}^{total}$+ $\mathcal{T}_3$}; 
\node (6b) [draw, align=center,right of=6a,xshift=2cm] {Target adaptation\\
\& fine-tuning};
  \node (6c) [draw, io, align=center,right of=6b,xshift=2cm] { Final Model};

\draw [arrow] (0a) -- (0b);
\draw [arrow] (0b) -- (0c);
\draw [arrow] (1a) -- (1b);
\draw [arrow] (1b) -- (1c);
\draw [arrow] (2a) -- (2b);
\draw [arrow] (2b) -- (2c);
\draw [arrow] (3a) -- (3b);
\draw [arrow] (3b) -- (3c);
\draw [arrow] (4a) -- (4b);
\draw [arrow] (4b) -- (4c);
\draw [arrow] (5a) -- (5b);
\draw [arrow] (5b) -- (5c);
\draw [arrow] (6a) -- (6b);
\draw [arrow] (6b) -- (6c);

\draw [arrow] (0c) -- (1b);
\draw [arrow] (1c) -- (2b);
\draw [arrow] (2c) -- (3b);
\draw [arrow] (3c) -- (4b);
\draw [arrow] (4c) -- (5b);
\draw [arrow] (5c) -- (6b);

\end{tikzpicture}
 \caption{Reiterative training approach assuming three target domains $\mathcal{T}_1,\mathcal{T}_2$, and $\mathcal{T}_3$ (in that order of closeness to the source domain) and two reiteration steps (i.e., $K^*=2$). Model and pseudo-source $\hat{\mathcal{S}}^{total}$ is updated continuously. }
 \label{figureReiterativeLearning}
\end{figure}
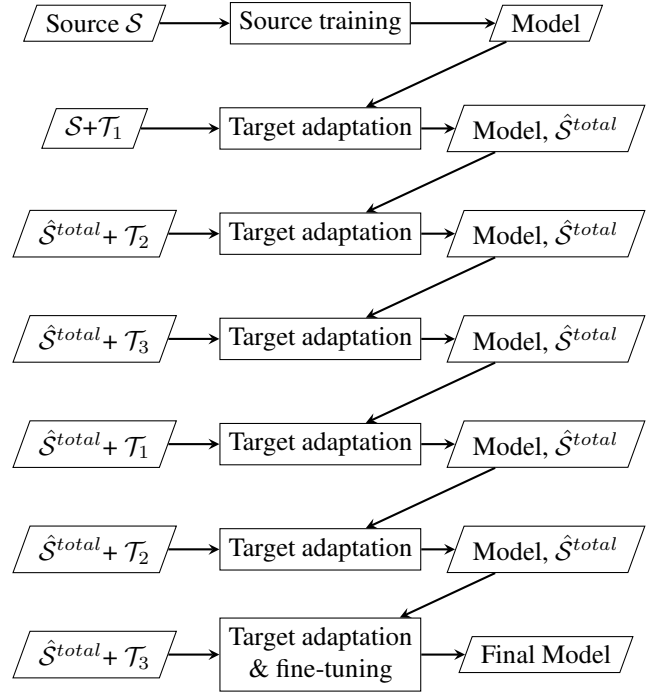

\begin{algorithm}
\small
\SetAlgoLined
\LinesNumbered
\SetCommentSty{mycommfont}
\SetKwInOut{Require}{require}
\SetKwInOut{Output}{output}
\SetKwComment{Comment}{$\triangleright$\ }{}
\Require{number of target domains $N$, classes $n_c$}
\Require{source dataset $\mathcal{S}$; target dataset $\mathcal{T} = \{\mathcal{T}_j\}^{N}_{j=1}$}
\Require{hyper-parameters $B$, $\tau$, $K$, $K^*$, $K', \lambda_{edge}, \lambda_{node}, \lambda_{adv}$}
\Require{networks $F$, $D$, $G_{mlp}$, $f_{edge}$, $f_{node}$ with parameters $\theta, \psi, \phi, \varphi, \varphi'$, respectively. The $f_{edge}$ and $f_{node}$ form the $G_{gnn}$.}

\nonl\underline{\textbf{Step 1}: \textit{Training on the source dataset}}\\
 \While{$\ell_{ce}$ has not converged}{
    ($\mathbf{x}_{s,i}, y_{s,i})^{B_s}_{i=1} \sim \mathcal{S}$\\
    update $\theta$, $\phi$ by $\Min_{\theta, \phi} \ell^{mlp}_{ce}$\\
 }
 \nonl\underline{\textbf{Step 2}: \textit{Curriculum learning}}\\
\DontPrintSemicolon $\hat{\mathcal{S}}^{total} \leftarrow \mathcal{S}$ \Comment*{Pseudo-source}     
 \DontPrintSemicolon $Q \leftarrow N$ \\
 \For{$k^* \; \text{in} \; (1 : K^*)$\Comment*{Reiterations}}{
 	$\hat{\mathcal{T}}^{0} \leftarrow \{\mathcal{T}_j\}^{N}_{j=1}$ \\
	 \For{$q$ $\text{in}$ (0 : $Q - 1$)}{
	    \DontPrintSemicolon 
	    \DontPrintSemicolon $\mathcal{H} \leftarrow \{\}$ \\
	    \nonl\underline{\textbf{Stage 1}: \textit{Domain selection stage}}\\
	    \For{$\mathcal{T}_j \; \text{in} \; \hat{\mathcal{T}}^{q}$}{
	        compute $H(\mathcal{T}_j)$ \\
	        $\mathcal{H} \leftarrow \mathcal{H} \; || \; H(\mathcal{T}_j)$ 
	    }
	    $\mathbb{D}^{q} \leftarrow \Argmin_{j} \mathcal{H}$\\ 
	    \nonl\underline{\textbf{Stage 2}: \textit{Adaptation stage}}\\
	    \For{$k \; \text{in} \; (1 : \frac{K}{K^*})$\Comment*{Adaptation iterations}}{  
	        $\hat{\mathcal{B}}^{q}_{s} \leftarrow (\mathbf{x}_{s,i}, y_{s,i})^{B_s}_{i=1} \sim \hat{\mathcal{S}}^{q}$ \\
	        $\hat{\mathcal{B}}^{q}_{t} \leftarrow (\mathbf{x}_{t,i})^{B_t}_{i=1} \sim \mathcal{T}_{\mathbb{D}^{q}}$\\
	        $\hat{y} \leftarrow \texttt{softmax} (G_{mlp}(F(\mathbf{x})))$ \\
	        $\bar{y} \leftarrow \texttt{softmax} (G_{gnn}(F(\mathbf{x})))$\\
	        $\hat{d} \leftarrow \texttt{sigmoid} (D(F(\mathbf{x})))$\\
	        update $\psi$ by $\Min_{\psi} \lambda_{adv} \ell_{adv}$\\
	        update $\theta$, $\phi$ by $\Min_{\theta, \phi} \ell^{mlp}_{ce} - \lambda_{adv} \ell_{adv}$\\
	        update $\theta$, $\varphi, \varphi'$ by $\Min_{\theta, \varphi, \varphi'} \lambda_{edge} \ell^{edge}_{bce} + \lambda_{node} \ell^{node}_{ce}$\\
	    }
	    \nonl\underline{\textbf{Stage 3}: \textit{Pseudo-labeling stage}}\\
	    $\mathcal{D}^{\mathbb{D}^{q}}_{t} \leftarrow \{\}$ 
	    \For{$\mathbf{x}_{t,j} \in \mathcal{T}_{\mathbb{D}^{q}}$}{
	        $w_j \leftarrow \Max_{c \in n_c} p(\bar{y}_{t,j} = c | \mathbf{x}_{t,j})$\\
	        \If{$w_j > \tau$}{
	            $\mathcal{D}^{\mathbb{D}^{q}}_{t} \leftarrow \mathcal{D}^{\mathbb{D}^{q}}_{t} || \{(\mathbf{x}_{t,j}, \ArgMax_{c \in n_c}p(\bar{y}_{t,j} = c | \mathbf{x}_{t,j})) \}$ 
	        }
	    }
	    $\hat{\mathcal{S}}^{total} \leftarrow \hat{\mathcal{S}}^{total} \cup \mathcal{D}^{\mathbb{D}^{q}}_{t}$ \Comment*{Pseudo-source}
	    $\hat{\mathcal{T}}^{q+1} = \hat{\mathcal{T}}^{q} \; \text{\textbackslash } \; \mathcal{T}_{\mathbb{D}^{q}}$\\
	 }
 }
 \nonl{\textbf{Step 3}: \textit{Fine-tuning on the pseudo-source dataset using $K'$ iterations}}\\
 
 \caption{Proposed training procedure}
 \label{algoRdcgct}
\end{algorithm}

\section{Proposed Method}
\label{sectionProposedMethod}
For the proposed multi-target domain adaptation, we are provided with
a source dataset $\mathcal{S}$ containing $n_s$ labeled samples $(\mathbf{x}_{s,i}, y_{s,i})^{n_s}_{i=1}$ and $N$ target datasets $\mathcal{T} = \{\mathcal{T}_j\}^{N}_{j=1}$. The underlying data distributions of the source and the $N$ target domains are different from
each other. However, they share the same label space and for each target sample their domain is known. Our goal is to learn predictor for the $N$ target domains by using the labeled data $\mathcal{S}$ and the unlabeled data $\mathcal{T}$.

\subsection{Network structure}
\label{sectionNetworkArchitecture}
While any backbone feature extractor can be used, we use a Transformer-based feature extractor network $F$ to extract the features from the input image, i.e., given an input $\mathbf{x}$, it generates the output $\mathbf{f}=F(\mathbf{x})$. Following this, features are fed to a MLP classifier $G_{mlp}$. Features are also fed to 
an edge-based network $f_{edge}$, the output of which is fed to $f_{node}$, a node classifier. The $f_{edge}$ and $f_{node}$ together form the $G_{gnn}$. Additionally, a domain discriminator network $D$ is used.
The parameters of networks $F$, $D$, $G_{mlp}$, $f_{edge}$, $f_{node}$ are represented as $\theta, \psi, \phi, \varphi, \varphi'$, respectively. 
The network structure is shown in Table \ref{tableNetworkArchitecture}.

\subsection{Transformer-based feature extractor}
\label{sectionViTFeatureExtractor}
We adopt the Transformer-based feature extractor from TransUNet \cite{chen2021transunet} as a pre-trained feature extractor $F$. Instead of using a pure Transformer, a CNN-Transformer hybrid model is used. CNN is first applied to the input image to generate feature map and subsequently patch embedding is applied to
1 $\times$ 1 patches extracted from the CNN feature map to compute the input of the Transformer. Such hybrid CNN-Transformer encoder has shown good performance both in image segmentation \cite{chen2021transunet} and domain adaptation \cite{yang2021transformer}. For the CNN, we specifically employ ResNet-50 model \cite{he2016deep}. The feature extractor backbone is pre-trained on ImageNet dataset \cite{deng2009imagenet}.

\subsection{Training on the source dataset}
\label{sectionSourcePretraining}
Before adapting on the targets $\mathcal{T}$, the source labeled samples $(\mathbf{x}_{s,i}, y_{s,i})^{n_s}_{i=1}$ are used to train the model $F$ and $G_{mlp}$, thus updating the parameters $\theta$ and $\phi$, respectively. While $F$ consists of ResNet-Transformer model as described in  Section \ref{sectionViTFeatureExtractor}, the 
$G_{mlp}$ is a fully connected output layer consisting of $n_c$ logits.
Cross-entropy loss $\ell^{mlp}_{ce}$ computed from the source samples is used for training and the target samples are not used at this stage.

\subsection{Pseudo-source}
\label{sectionPseudoSource}
At each step of the target adaptation, we treat a  selected set of samples $\hat{\mathcal{S}}^{total}$ as the samples for which we are  confident about their labels. These samples are called pseudo-source ($\hat{\mathcal{S}}^{total}$) and is initialized with source samples
$\mathcal{S}$ after the step in Section \ref{sectionSourcePretraining}. However target samples are slowly added to it, as explained in  the following subsections. 

\subsection{Target adaptation}
\label{sectionTargetAdaptation}
In this step the model performs feature adaptation on the target domains, one domain at a time, starting from the easiest domain and gradually moving to the more difficult ones.
The easiest domain is defined as the domain closest to the source domain, which is measured by the uncertainty ($H(\mathcal{T}_j)$ for domain $\mathcal{T}_j$) in the target predictions 
with the source-trained model \cite{roy2021curriculum,gal2016uncertainty}.
\par
Once one target domain to be processed is fixed, adaptation is performed for $K$ iterations.  In each iteration, some source samples ($\hat{\mathcal{B}}^{q}_{s}$) and some  target samples ($\hat{\mathcal{B}}^{q}_{t}$) are drawn to form a minibatch.
Each minibatch of images is fed to the feature extractor to obtain features corresponding to them and then fed to the $G_{mlp}$ and $G_{gnn}$.
The  $G_{gnn}$ consists of the edge network $f_{edge}$ and the  node classifier $f_{node}$.
The $G_{mlp}$ does not aggregate features from different samples, rather predict based on only the sample of interest. On the other hand GNN-based classifier aggregates the features of the samples in the
batch. In other words, prediction from $G_{gnn}$ not only accounts for a sample $\mathbf{x}$, rather also for the other samples in the minibatch. While this leads to potentially better context-aware learning paradigm, naively using it may lead
to noisy feature degrading the classification performance. The MLP classifier and the GNN classifier capture different aspects, thus they are further exploited to provide feedback to each other, similar to co-teaching \cite{han2018co}. 
The output from the MLP head is obtained as:
\begin{equation}
\hat{y} \leftarrow \texttt{softmax} (G_{mlp}(F(\mathbf{x})))
\end{equation}
Similarly, the output from the GNN head is obtained as:
\begin{equation}
\bar{y} \leftarrow \texttt{softmax} (G_{gnn}(F(\mathbf{x})))
\end{equation}
The $G_{mlp}$ and $f_{node}$ of $G_{gnn}$  is trained with cross-entropy loss $\ell^{mlp}_{ce}$ and $\ell^{edge}_{ce}$, respectively, computed over the source/pseudo-source samples.
\par
\textbf{Feedback from MLP to GNN:} For edge network $f_{edge}$ to learn the pairwise similarity between samples, a target matrix $\hat{\mathcal{A}}^{tar}$ is formed such that an element  $\hat{a}^{tar}$ in
$\hat{\mathcal{A}}^{tar}$ is $1$ if the labels of $i$-th and $j$-th sample are same. While label information is known from the  source samples ($\hat{\mathcal{B}}^{q}_{s}$), they are not known for the 
target samples ($\hat{\mathcal{B}}^{q}_{t}$). Thus, pseudo-labels of target samples are formed using prediction of $G_{mlp}$. In this way $G_{mlp}$ teaches $G_{gnn}$. A (binary-cross-entropy) edge
loss between elements of affinity matrix $\hat{\mathcal{A}}$ (produced by $f_{edge}$) and the target matrix $\hat{\mathcal{A}}^{tar}$  is computed as
$\ell^{edge}_{bce}$.
\par
\textbf{Feedback from GNN to MLP:} At the end of processing each target domain, a score $w_j$ is assigned to each sample $\mathbf{x}_{t,j}$ in the target domain as:
\begin{equation}
w_j \leftarrow \Max_{c \in n_c} p(\bar{y}_{t,j} = c | \mathbf{x}_{t,j})
\end{equation}
$w_j$ indicates the confidence of prediction of $G_{gnn}$ for the sample $j$. If $w_j$ is greater than a score $\tau$, then this sample is appended to the list of pseudo-labeled samples $\hat{\mathcal{S}}^{total}$ that are used while processing the subsequent 
target domains. In this way, the $G_{gnn}$ creates a set of context-aware pseudo-samples after processing each target domain, thus enabling more effective learning of $G_{mlp}$.
\par
\textbf{Domain discriminator:} In addition to the MLP and GNN based networks, a domain discriminator $D$ is used to predict the domain of the samples \cite{ganin2015unsupervised,long2017conditional}. The output from this is obtained as:
\begin{equation}
\hat{d} \leftarrow \texttt{sigmoid} (D(F(\mathbf{x})))
\end{equation}
The domain discriminator in conjunction with a gradient reversal layer (GRL) is trained with an adversarial loss $\ell_{adv}$. This further pushes the feature extractor  to learn features  such that the source and different target features
appear as if they are coming from the same distribution.
\par
The target adaptation processing assuming a single target is shown in 
Figure \ref{figureSingleTargetAdaptation}.

\subsection{Reiterative target adaptation}
\label{sectionReiterativeAdaptation}
The target adaptation, as explained in Section \ref{sectionTargetAdaptation}   makes only one pass through the easiest to the most difficult domain (with $K$ training iterations for each domain). The model will not be sufficiently adapted to the target domains if  the chosen value of $K$ is small. However, if value of $K$ is big, the softmax values produced by the model may be large even for the samples for which the model is not confident \cite{guo2017calibration}. This may cause erroneous samples selected in pseudo-source $\hat{\mathcal{S}}^{total}$.  To circumnavigate this problem, we propose to make multiple ($K^*$) passes (we call it reiteration) through each domain. 
\par
In more details, if there are only three target domains $\mathcal{T}_1$, $\mathcal{T}_2$, and $\mathcal{T}_3$ (in that order of difficulty), the target adaption in Section
\ref{sectionTargetAdaptation}  processes it only once in the order
$\mathcal{T}_1$, $\mathcal{T}_2$, and $\mathcal{T}_3$. Instead, using the reiterative strategy, if $K^*=2$,  proposed method processes the target domains in the order $\mathcal{T}_1$, $\mathcal{T}_2$, $\mathcal{T}_3$, $\mathcal{T}_1$, $\mathcal{T}_2$, and $\mathcal{T}_3$. Reiterative approach
is further illustrated in Figure \ref{figureReiterativeLearning}.
\par
We also propose to scale down the number of training iterations (per reiteration) by a factor of $K^*$. In this way, proposed method with $K^*$ reiteration passes
uses the same number of iterations as original D-CGCT with $K$ adaptation iterations per target domain.
By only iterating $K/K^*$ times per target domain in a reiteration, the target adaptation process does not push itself too hard, thus avoiding to produce high softmax values for low-confident samples. On the other hand, with increasing reiterative passes, the adaptation process  unfolds slowly, however in a more effective manner.

\subsection{Emphasizing source samples}
\label{sectionMoreSourceSamples}
We furthermore postulate that more source samples in a minibatch can lead to more effective target adaptation. In other words, instead of setting $B_s=B_t=B$, we propose to set $B_s>B_t$. Such a
modification does not affect the number of training iterations. Moreover, $B_s+B_t$ can be kept fixed by simply decreasing $B_t$ while increasing $B_s$. 
\par
The proposed method is detailed in Algorithm \ref{algoRdcgct}.

\subsection{Using trained model for inference}
\label{sectionUsingModelInference}
Once the proposed MLP-GNN dual head model is trained, either head can be used during inference to determine the class of a target test sample. However, as noted in \cite{roy2021curriculum}, the GNN head requires
mini-batch of samples, that may not be available during real-world inference. Hence, once trained, it is more practical to use the MLP head for inference.

\section{Experimental Validation}
\label{sectionExperimentalResult}
\subsection{Datasets}
We conducted experiments on standard three domain adaptation datasets.
\par
\textbf{Office-Home} dataset contains 4 domains (Art, Clipart, Product, Real) and 65 classes \cite{venkateswara2017deep}.
\par
\textbf{Office-31} dataset contains 3 domains (Amazon, DSLR, Webcam) and 31 classes \cite{saenko2010adapting}.
\par
\textbf{DomainNet} dataset contains 6 domains and 345 classes \cite{peng2019moment}. Compared to the other two datasets, this one is very large scale containing 0.6 million images. The 6 domains are Real (R), Painting (P), Sketch (S), Clipart (C), Infograph (I), Quickdraw (Q).

\subsection{Evaluation protocol and settings}
Like previous works on MTDA \cite{roy2021curriculum}, we use the classification accuracy to evaluate the performance of proposed method. 
The performance for a given source is given by setting the remaining domains as target domains and averaging the accuracy on all the target domains.
\par
For sake of fairness of comparison, we use the the same set of hyperparameters as reported in \cite{roy2021curriculum}.  In addition to D-CGCT \cite{roy2021curriculum}, comparison is  shown to MT-MTDA \cite{nguyen2021unsupervised} and 
Conditional Adversarial Domain Adaptation (CDAN) \cite{long2017conditional} along with domain-aware curriculum learning (DCL), i.e., CDAN+DCL, as shown in \cite{roy2021curriculum}. For DomainNet, comparison is also provided to HGAN \cite{yang2020heterogeneous}.

\begin{table}[t]
\centering
\begin{tabular}{l|l|l|l|l} 
 \hline
\textbf{$k$} & \textbf{Corr.} & \textbf{Incorr.} & \textbf{$\Delta$ corr.} & \textbf{$\Delta$ incorr.}\\ 
\hline
1   & 2518 & 338 & - & -  \\ 
500   & 2798 & 389 & 280 & 51  \\ 
1000   & 2868 & 445 & 70 & 56  \\
\end{tabular}
\caption{Variation of correct and incorrect pseudo-samples as adaptation iteration ($k$) progresses. The last two columns indicate the change in number of correct and incorrect pseudo-samples. The result is shown Office-Home dataset for Product as target, using Art as source.}
\label{tablePseudoSampleNumberSourceArtTargetProduct}
\end{table}

\begin{table}[t]
\centering
\begin{tabular}{l|l|l|l|l|l} 
\textbf{$K^*$} & 1 & 3  & 5 & 10 & 20 \\ 
\hline
\textbf{Acc.} & 70.5  & 72.4  & 73.4 & \bf{73.6} & 72.8  \\
\end{tabular}
\caption{The performance variation w.r.t. reiteration numbers $(K^*)$ on Office-Home dataset, source - Art, target - rest. The target accuracy is averaged over all the target
 domains. We observe that as $K^*$ is increased, target classification accuracy either improves or remains similar. The best result is obtained for $K^*=10$. }
\label{tableVariationWithDifferentReiteration}
\end{table}

\begin{table}[t]
\centering
\begin{tabular}{l|l|l|l|l} 
 \hline
\textbf{$k^*$} & \textbf{Real} & \textbf{Product} & \textbf{Clipart} & \textbf{Target Acc.}\\ 
\hline
1   & 2788 & 2912 & 2058 & 65.6  \\ 
2   & 769 & 810 & 843 & 70.0   \\ 
3   & 280 & 255 & 420 & 71.3  \\ 
4   & 137 & 162 & 250 & 72.2  \\ 
5   & 68 & 82 & 160 & 72.8  \\ 
6   & 53 & 59 & 106 & 73.0  \\ 
7   & 35 & 30 & 80 & 73.4  \\
8   & 26 & 20 & 55 & 73.4  \\
9   & 20 & 17 & 44 & 73.5  \\
10   & 11 & 11 &  32 & 73.6  \\
\end{tabular}
\caption{Column 2-4 indicates the number of pseudo-samples selected from the target domains after each re-iteration $k^*$ (Office-Home dataset, taking Art as source). Note that total reiteration is fixed as $K^*=10$. Final column indicates the average target classification accuracy after each reiteration}
\label{tableOfficeHomePseudoSampleSelection}
\end{table}

\begin{table}[t]
\centering
\begin{tabular}{l|l|l|l} 
\textbf{$B_s,B_t$} & 32,32 & 48,16 & 48,32 \\
 \hline
\textbf{Target Acc.} & 73.6 & 73.7 & 75.1  \\ 
\hline
\end{tabular}
\caption{The performance variation w.r.t. source and target samples in a minibatch $(B_s,B_t)$ on Office-Home dataset, source - Art, target - rest. The target accuracy is averaged over all the target
 domains. }
\label{tableVariationSourceTargetSamples}
\end{table}

\begin{table}[t]
\centering
\begin{tabular}{l|l|l|l} 
 \hline
$K^*$,$B_s$,$B_t$ &  \textbf{Real} & \textbf{Product}& \textbf{Clipart}  \\ 
\hline 
1,32,32   & 2969 &  3047 & 1971 \\
10,48,16   & 3537 &  3480 & 2580 \\
\end{tabular}
\caption{Correct pseudo-sample at last iteration. Dataset: Office-Home, Source: Art, target: rest.}
\label{tablePseudoSampleAccuracyComparison}
\end{table}

\begin{table}[t]
\centering
\begin{tabular}{l|l|l|l|l} 
 \hline
$k^*$ & $B_s$ & $B_t$ & \textbf{Backbone} & \textbf{Acc.} \\ 
\hline
1   & 32 & 32 & ResNet-50 & 70.5  \\ 
10   & 32 & 32 & ResNet-50 & 73.6  \\
10   & 48 & 16 & ResNet-50 & 73.7  \\
10   & 48 & 16 & Transformer & 80.8  \\ 
\end{tabular}
\caption{ Variation of 4 components of the proposed method ($k^*$, $B_s$, $B_t$,
and backbone) on Office-Home dataset, source: Art, target: rest.}
\label{tableResultDifferentComponentVariation}
\end{table}

\begin{table}[t]
\centering
\begin{tabular}{l|l|l|l|l} 
 \hline
\textbf{Method} & \textbf{Art} & \textbf{Clip.} & \textbf{Prod.} & \textbf{Real}   \\ 
\hline
D-CGCT (ResNet-18)   & 61.4 & 60.7 & 57.3 & 63.8  \\
Proposed (ResNet-18)  & \bf{64.9} & \bf{65.4} & \bf{61.8} & \bf{66.7}  \\
\hline
D-CGCT (ResNet-50)   & 70.5 & 71.6 & 66.0 & 71.2 \\
Proposed (ResNet-50)   & \bf{73.7} & \bf{76.0} & \bf{68.7} & \bf{72.7} \\
\hline
D-CGCT (Transformer)   & 77.0 & 78.5 & 77.9 & \bf{80.9} \\
Proposed (Transformer)   & \bf{80.8} & \bf{81.8} & \bf{80.7} & 78.8 \\
\end{tabular}
\caption{The performance comparison on the Office-Home dataset when using different architectures as backbone.}
\label{tableOfficeHomeWithDifferentArchitectures}
\end{table}

\begin{table}[t]
\centering
\begin{tabular}{l|l|l|l|l} 
 \hline
\textbf{Method} & \textbf{Art} & \textbf{Clipart} & \textbf{Product}& \textbf{Real}  \\ 
\hline 
MT-MTDA   & 64.6 & 66.4 & 59.2 & 67.1 \\
CDAN+DCL   & 63.0 & 66.3 & 60.0 & 67.0 \\
D-CGCT   & 70.5 & 71.6 & 66.0 & 71.2 \\
Proposed   & \bf{80.8} & \bf{81.8} & \bf{80.7} & \bf{78.8} \\
\end{tabular}
\caption{The performance comparison on the Office-Home dataset. The classification accuracy is reported for each
source and the rest set as target, with each source domain being indicated in the columns.}
\label{tableOfficeHome}
\end{table}

\begin{table}[t]
\centering
\begin{tabular}{l|l|l|l} 
 \hline
\textbf{Method} & \textbf{Amazon} & \textbf{DSLR} & \textbf{Webcam} \\ 
\hline
MT-MTDA   & 87.9 & 83.7 & 84.0  \\
CDAN+DCL   & 92.6 & 82.5 & 84.7  \\
D-CGCT   & 93.4 & 86.0 & 87.1 \\
Proposed & \bf{93.4}  & \bf{89.4}  & \bf{89.9}   \\
\end{tabular}
\caption{The performance comparison on the Office-31 dataset. The classification accuracy is reported for each
source and the rest set as target, with each source domain being indicated in the columns.}
\label{tableOffice31}
\end{table}

\begin{table*}[t]
\centering
\begin{tabular}{l|l|l|l|l|l|l|l|l} 
 \hline
\textbf{Method} & \textbf{R$\rightarrow$S} & \textbf{R$\rightarrow$C} & \textbf{R$\rightarrow$I} & \textbf{R$\rightarrow$P} &  \textbf{P$\rightarrow$S} & \textbf{P$\rightarrow$R}  & \textbf{P$\rightarrow$C}  & \textbf{P$\rightarrow$I}   \\ 
\hline
HGAN   & 34.3 & 43.2 & 17.8 & 43.4 & 35.7 & 52.3 & 35.9 & 15.6  \\
CDAN+DCL   & 45.2 & 58.0 & 23.7 & 54.0 & 45.0 & \bf{61.5} & 50.7 & 20.3  \\
D-CGCT   & 48.4 & 59.6 & 25.3 & 55.6 & 45.3 & 58.2 & \bf{51.0} & 21.7  \\
Proposed &  \bf{58.6} & \bf{67.7}  & \bf{30.9}  & \bf{61.0}  & \bf{47.5}  & 55.9  & 48.6 & \bf{24.2}   \\
\end{tabular}
\caption{The performance comparison on the DomainNet dataset.  Domains are Real (R), Painting (P), Sketch (S), Clipart (C), and Infograph (I). R$->$S indices Real as source and Sketch as target, similarly for other cases.}
\label{tableDomainNet}
\end{table*}

\subsection{Analyses on Office-Home dataset}
\label{SectionParameterAnalyses}
We perform several analyses and ablation studies on the Office-Home dataset.
These experiments simply use ResNet-50 (not Transformer) backbone as in \cite{roy2021curriculum}, unless otherwise stated.
\par
We hypothesized in Section \ref{sectionReiterativeAdaptation} that
attempting to adapt model on a target for too many iterations (big $K$) may instead produce large softmax values even for samples for which model is not confident, thus yielding unreliable pseudo-samples. To verify this, we show the progress of pseudo-sample selection for Art as source and Product as target in Table \ref{tablePseudoSampleNumberSourceArtTargetProduct}. It is evident that for the first few hundred iterations, more correct samples are added. However as training is prolonged, more and more incorrect samples exceed the softmax threshold and are added to pseudo-source. This validates our hypothesis that attempting to prolong training may impact the adaptation process negatively. 
\par
Taking Art as source and rest as target, we vary the number of reiteration ($K^*$), as explained in Section \ref{sectionReiterativeAdaptation}. In this experiment, we use $B_s=32, B_t=32$. Here, $K^*=1$ is equivalent to using D-CGCT. We observe that the target accuracy improves as  $K^*$ increases. Best target accuracy is obtained at $K^*=10$. Following \cite{roy2021curriculum}, we used $K=10000$. Thus, $K^*=10$ implies decomposing the adaptation process into 10 reiterations of $\frac{K}{K^*}=1000$ iterations each. In the rest of the experiments related to Office-Home dataset, $K^*=10$ is used.
\par
After each reiteration, some new target samples are added to the pseudo-source $\hat{\mathcal{S}}^{total}$. 
Taking Art as source and rest as target, we show this process in Table
\ref{tableOfficeHomePseudoSampleSelection} for $K^*=10$.
We observe that the reiterative process allows to gradually ingest the pseudo-samples, more at the  first reiteration and gradually less in the subsequent reiterations. E.g., for the target domain Real, 2788 samples are added to pseudo-source at the first reiteration, followed by 769 and 280 new samples added to the pseudo-source in the following two reiterations and so on.
A gradual improvement in average target classification accuracy is observed over the reiterations. 
\par
We hypothesized in Section \ref{sectionMoreSourceSamples} that using more source samples in the training minibatch may improve the target adaptation. To examine this, we vary value of 
$B_s$ and $B_t$, as shown in Table \ref{tableVariationSourceTargetSamples} taking Art as source and rest as target.
Among the three compared settings, $B_s=48$, $B_t=32$ obtains the best result. Even with $B_s=48$ and $B_t=16$, improvement is observed over $B_s=B_t=32$. Though $B_s=48$, $B_t=32$ is better, we use $B_s=48$, $B_t=16$ combination for subsequent experiments, to keep 
$B_s+B_t$ value same as in \cite{roy2021curriculum}.
\par
We further show in Table \ref{tablePseudoSampleAccuracyComparison} that the correctly chosen pseudo-samples increase after employing reiterative strategy and using more source samples in minibatch. 
\par
Using Transformer as the backbone feature extractor further improves the performance of the proposed method, as shown in Table \ref{tableResultDifferentComponentVariation}.
\par
Finally we show that the proposed reiterative strategy outperforms
D-CGCT \cite{roy2021curriculum} in most cases irrespective of the feature extractor backbone. This is shown on all 4 domains 
(Art, Clipart, Product, Real) as source, while treating other three as target. 
Performance improvement is consistent for ResNet-18, ResNet-50, and Transformer (except one case), as shown in Table
\ref{tableOfficeHomeWithDifferentArchitectures}.
\par
Henceforth, the proposed method is shown 
with Transformer-based feature extractor, as detailed in Section \ref{sectionViTFeatureExtractor}.
Accuracy reported for D-CGCT is however from \cite{roy2021curriculum}, with architecture of their choice.

\subsection{Quantitative comparisons}
\label{sectionQuantitativeComparison}
\textbf{Office-Home:}
The proposed method  outperforms D-CGCT (and other compared methods) for all four domains. The performance improvement is quite prominent,  Art (10.3$\%$), Clipart (10.2$\%$), Product (14.7$\%$), and Real (7.6$\%$). This may be attributed to both use of Transformer as backbone feature extractor and the reiterative training strategy.
The performance improvement is almost uniform for all domains, except for the Real domain. On average it outperforms D-CGCT by 10.7$\%$.  The quantitative result
 of Office-Home dataset is shown in Table \ref{tableOfficeHome}. 

\par
\textbf{Office-31:} Originally in \cite{roy2021curriculum}, target adaptation iterations $K$ is set as 3000 for Office-31 dataset. We set number of reiterations $K^*=3$ to keep $\frac{K}{K^*}=1000$, as in Office-Home.
\par
 The proposed method  outperforms D-CGCT for DSLR (3.4$\%$) and Webcam (2.8$\%$) and obtains similar performance for Amazon. Remarkably, proposed method shows the least improvement for Amazon as source, which has the most number of images in Office-31 dataset. This indicates a possibility that the improvement brought by the proposed method diminishes if there are abundant images in the labeled source domain, which is often not the case in practice. 
The quantitative result of the Office-31 dataset  is shown in Table \ref{tableOffice31}. 
\par
\textbf{DomainNet:}  Like other two datasets, we chose $K^*$ such that $\frac{K}{K^*}=1000$.
The quantitative result for the different source $\rightarrow$ target combinations are
 shown in Table \ref{tableDomainNet}, following the format in \cite{yang2020heterogeneous}. Performance improves over D-CGCT in most cases, e.g., R$\rightarrow$S (10.2$\%$), R$\rightarrow$C (8.1$\%$), R$\rightarrow$I (5.6$\%$), R$\rightarrow$P (5.4$\%$), P$\rightarrow$S (2.2$\%$) and  P$\rightarrow$I (2.5$\%$). 
 
\section{Limitations}
\label{sectionLimitation}
The proposed method requires data from the source domain while adapting the model to target.
This may potentially limit its practical application and needs to be addressed in the future by modifying the proposed method for source-free domain adaptation \cite{kundu2021generalize}. The method does not account for possible presence of open sets \cite{ma2021active,xia2021adaptive}.  While use of a heavier backbone (Transformer) may also be considered a limitation of the proposed method, it is shown in  
Section \ref{SectionParameterAnalyses} that proposed method outperforms other methods even for lighter architectures.

\section{Conclusion}
\label{sectionConclusion}
While single-source single-target domain adaptation has been long studied in the literature,
its practical applications are limited. Thus domain adaptation is moving beyond this simple setting towards more practical and complex settings, e.g., multi-target adaptation. Our work takes forward multi-target adaptation by exploiting Transformer along with graph neural network. The proposed reiterative adaptation strategy enhances the target adaptation performance by enabling the network to select more accurate pseudo-samples. The proposed method consistently improves result for Office-Home and Office-31 datasets and almost always for the DomainNet dataset. In the future we plan to extend the method for multi-source muti-target adaptation and source-free adaptation.

\bibliographystyle{ieeetr}
\bibliography{reiterativeDA}

\end{document}